\documentclass[10pt,twocolumn,letterpaper]{article}

\usepackage{cvpr}
\usepackage{times}
\usepackage{epsfig}
\usepackage{graphicx}
\usepackage{amsmath}
\usepackage{amssymb}
\usepackage[english]{babel}
\usepackage[utf8]{inputenc}
\usepackage{listings}
\usepackage{svg}

\usepackage{xcolor}

\definecolor{codegreen}{rgb}{0,0.6,0}
\definecolor{codegray}{rgb}{0.5,0.5,0.5}
\definecolor{codepurple}{rgb}{0.58,0,0.82}
\definecolor{backcolour}{rgb}{1,1,1}

\lstdefinestyle{mystyle}{
  backgroundcolor=\color{backcolour},   commentstyle=\color{codegreen},
  keywordstyle=\color{magenta},
  numberstyle=\tiny\color{codegray},
  stringstyle=\color{codepurple},
  basicstyle=\ttfamily\footnotesize,
  breakatwhitespace=false,         
  breaklines=true,                 
  captionpos=b,                    
  keepspaces=true,                 
  numbers=left,                    
  numbersep=5pt,                  
  showspaces=false,                
  showstringspaces=false,
  showtabs=false,                  
  tabsize=2
}

\graphicspath{ {images/} }

\usepackage[pagebackref=true,breaklinks=true,letterpaper=true,colorlinks,bookmarks=false]{hyperref}

\cvprfinalcopy 

\ifcvprfinal\fi
\begin{document}

\title{Self-attention aggregation network for video face representation and recognition}

\author{Ihor Protsenko$^{1,4}$, Taras Lehinevych$^{1,2,4}$, Dmytro Voitekh$^{4,5}$, Ihor Kroosh$^{3,4,5}$, Nick Hasty$^5$, Anthony Johnson$^5$ \\
 {$^1$~National University of Kyiv-Mohyla Academy}\\
 {$^2$~Institute of Software Systems of NAS of Ukraine}\\
 {$^3$~National Technical University of Ukraine
“Igor Sikorsky Kyiv Polytechnic Institute”}\\
 {$^4$~Proxet} \\
 {$^5$~GIPHY}\\
 {\tt\small i.protsenko@ukma.edu.ua, research@taraslehinevych.me, dmitry.voitekh@railsreactor.com,} \\
 {\tt\small ihor.kroosh@gmail.com, nick@giphy.com, anthony@giphy.com } \\
}

\maketitle

\begin{abstract}
  Models based on self-attention mechanisms have been successful in analyzing temporal data and have been widely used in the natural language domain. We propose a new model architecture for video face representation and recognition based on a self-attention mechanism. Our approach could be used for video with single and multiple identities. To the best of our knowledge, no one has explored the aggregation approaches that consider the video with multiple identities. The proposed approach utilizes existing models to get the face representation for each video frame, e.g., ArcFace and MobileFaceNet, and the aggregation module produces the aggregated face representation vector for video by taking into consideration the order of frames and their quality scores. We demonstrate empirical results on a public dataset for video face recognition called IJB-C to indicate that the self-attention aggregation network (SAAN) outperforms naive average pooling. Moreover, we introduce a new multi-identity video dataset based on the publicly available UMDFaces dataset and collected GIFs from Giphy. We show that SAAN is capable of producing a compact face representation for both single and multiple identities in a video. The dataset and source code will be publicly available. Code: \href{https://github.com/lehinevych/SAAN}{https://github.com/lehinevych/SAAN}

\end{abstract}

\section{Introduction}

Video face recognition has received increasing interest from the community in recent years \cite{crosswhite2018template, cui2013fusing, hu2014discriminative, li2013probabilistic, liu2014toward, mendez2013volume, parkhi2014compact, taigman2014deepface, schroff2015facenet}, partially due to the growing volume of video content. Compared to an image, a video sequence contains more information about the subject's faces, such as varying poses, expressions, motion, and illumination. The key challenge for video-based face recognition is how to effectively combine facial information available across different video frames to get an appropriate representation of the face in the video.

Deep Convolutional Neural Networks (DCNNs) trained on a large dataset have shown the ability to generate compact and discriminative face representations for images that are robust to pose variations, image quality, blur, and occlusions \cite{liu2017sphereface, schroff2015facenet, sun2015deeply, taigman2014deepface}. The naive approach is to represent video as a set of frames and use face features extracted by deep neural network for each frame \cite{schroff2015facenet, taigman2014deepface}. Then the subject's face in the video, called face track, is represented as an unordered set of vectors that allows maintaining the information across all frames. However, this is not computationally efficient for comparing two face tracks. It requires a fusion of matching results and comparison across all pairs of vectors between two face templates. Except for $O(n^2)$ complexity per match operation, face track requires $O(n)$ space per video face, where $n$ - the average number of video frames. That is why most methods aggregate feature vectors into fix-size feature representation for each face track \cite{liu2018dependency, rao2019learning, rao2017attention, sohn2017unsupervised, xie2018comparator, yang2017neural}. It allows constant-time matching computation instead of evaluating all powers. 

Initially, the aggregation challenge was considered when GIPHY has built its own open-source model called “Giphy Celebrity Detector”\cite{} to automatically annotate  GIFs featuring celebrities. This provides better search results for entertainment related queries, which is a vital aspect of GIPHY everyday traffic. Due to the fact that GIFs could be also represented as a set of frames we consider the feature aggregation networks. Unfortunately, the existing methods don't consider multi-identity face appearance in frames. For example, GIFs from popular shows contain more than one face.


In this paper, we propose an aggregation model based on self-attention, which can be applied for videos that contain single or multiple identities. Also, we create a synthetic dataset for multi-identity video face recognition.

\section{Related Work}
\subsection{Image Face Recognition}

Face recognition is an actively studied domain with significant achievements in identification and verification tasks \cite{wang2016face, wang2018cosface, zhang2016joint, liu2017sphereface} and a great part of that success is due to deep convolutional neural networks. Most approaches are focused on learning the embedding vector for face representation. The following works focus on exploring different loss functions to improve the feature representation. Both contrastive \cite{chopra2005learning, hadsell2006dimensionality} and triplet \cite{hoffer2015deep, wang2014learning} losses are usually used to increase the Euclidean margin for better feature embedding.
The center loss proposed in \cite{wen2016discriminative} learns centroids for features of each identity to reduce intra-class variance. A large margin softmax (L-softmax) \cite{liu2016large} adds angular constraints to each identity to improve feature discrimination, and angular softmax (A-softmax) \cite{liu2017sphereface} adds weights normalization for L-softmax. The ArcFace uses additive angular margin loss \cite{deng2019arcface}.

\subsection{Video Face Recognition} 
Existing methods are classified into the following categories: ones that exploit temporal dynamics and ones that treat video as  an orderless set of images. The first group of methods heavily relies on RNNs to account for the temporal dependencies in frame sequences. For example, RNN was employed for head pose estimation \cite{gu2017dynamic}, facial expression recognition \cite{graves2008facial}, and emotion recognition \cite{fan2016video, zhang2018spatial}. Many previous methods have considered the representation of the set of face images as probabilistic distribution \cite{arandjelovic2005face, shakhnarovich2002face}, n-order statistics \cite{lu2013image}, affine hulls \cite{yang2013face, hu2011sparse, cevikalp2010face},
SPD matrices \cite{huang2015log}, manifolds \cite{lee2003video, harandi2011graph, wang2008manifold}, etc. Then, the recognition is performed via similarity or distance measures. Other methods aim to train a supervised classifier on each image set or video to obtain correspondent representation \cite{parchami2017using}. These methods work well under constrained settings but usually struggle to handle the challenging unconstrained scenarios with significant appearance variations. The aggregation based models aim to fuse a set of deep feature vectors into a single vector. It was shown in \cite{best2014unconstrained} that aggregation of multiple face images increases the recognition performance of person identification. Compared to simple average pooling \cite{chen2018unconstrained, ding2017trunk, chen2015end, chowdhury2016one, deng2019arcface}, the recent works show promising results by incorporating the visual quality information on instance level via detection score \cite{ranjan2018crystal} or predicted quality scores \cite{yang2017neural, liu2017quality}. A Component-wise Feature Aggregation Network \cite{gong2019video} aggregates the feature vectors in each component separately by considering the prediction of their quality. The redundancy issue in the video frames is tackled in \cite{gong2019low}. However, none of these approaches consider the multi-identity video recognition setup.

\section{Aggregation networks}
In the following sections, we'll review proposed single and multi-identity aggregation architectures. To our knowledge, the aggregation of multiple identities has not been addressed so far. We consider a multi-head self-attention mechanism that has been successfully used to encode/decode sequence representation as in \cite{vaswani2017attention}, achieving superior results and better parallelization compared to recurrent encoder/decoder framework. It is used to achieve reweighting according to the context of the features. 

\begin{figure}
   \begin{center}
   \includegraphics[width=1.0\linewidth]{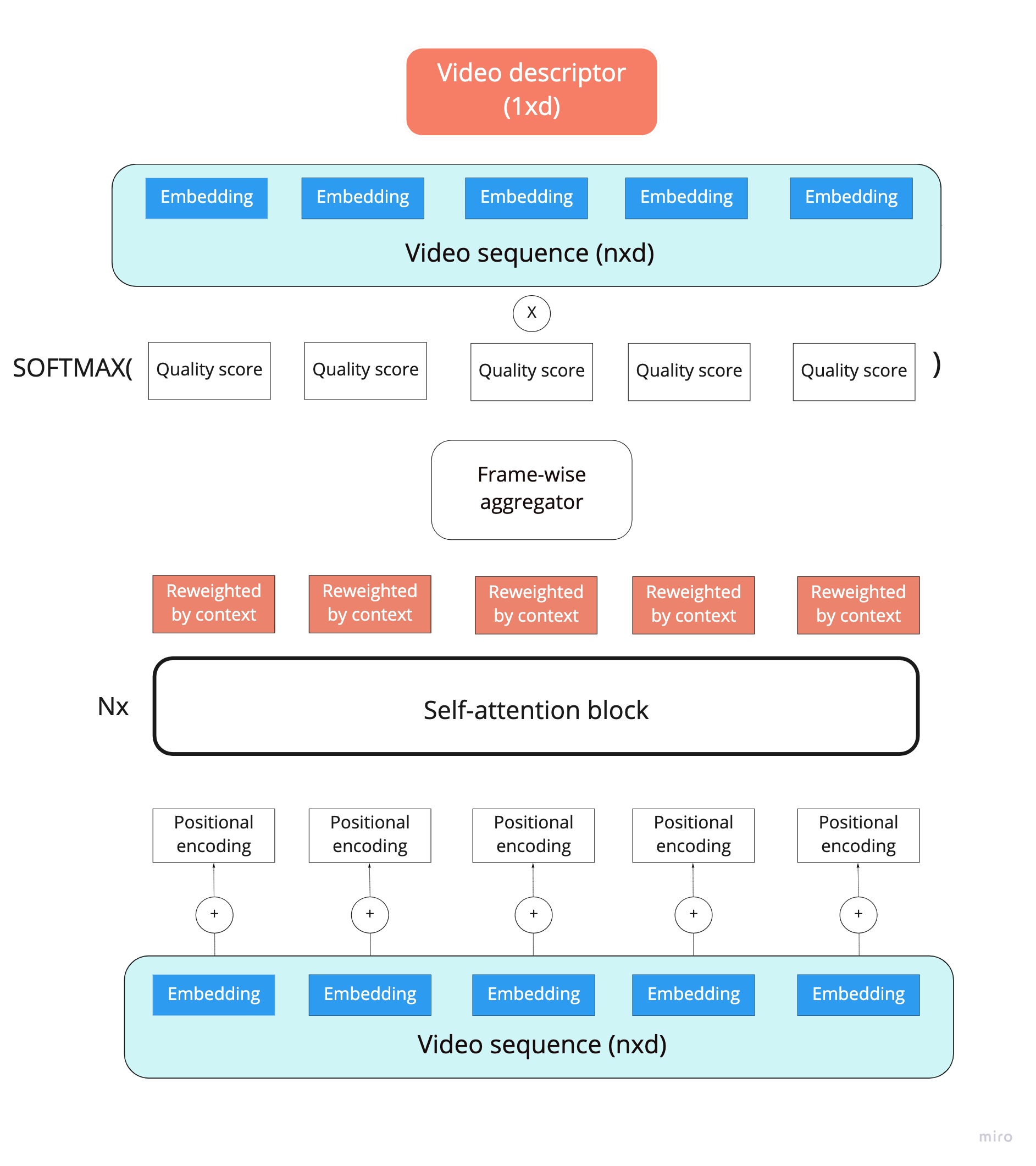}
   \end{center}
   \caption{Example of video sequence aggregation network using self-attention. $n$ is the number of faces in a face track, $d$ is a dimensionality of the embedding, $N$ - is the number of consecutive self-attention blocks of the encoder. Figure is read in a bottom-up fashion.}
   \label{fig:self_attention_arc}

\end{figure}
\subsection{Feature extractor}

To get a compact representation for each image in a given set, we use a publicly available \href{https://github.com/deepinsight/insightface}{ArcFace model}
\cite{deng2019arcface} pretrained on the MS1MV2 dataset \cite{guo2016ms}. It is considered to be state-of-the-art in a lot of public verification/identification benchmarks at the time.

Additionally, we experiment with a lightweight feature extractor, in particular, MobileFaceNet \cite{chen2018mobilefacenets}. The main purpose of that experiment is to investigate the aggregation results on features with a lower representational capacity.

\subsection{Single-identity aggregation}
\label{sident}
Given a set of ordered frames $\{x_1, x_2, x_3, x_4,\dots ,x_n\}$ (face track) and their corresponding features $\{f_1, f_2, f_3, f_4,\dots ,f_n\}$, aggregation should be performed to produce a single vector representation $\boldsymbol{r}$. To represent a face track from different perspectives and assign weight to a particular frame with respect to its context, we use a multi-head self-attention mechanism (Figure \ref{fig:self_attention_arc}). At first, $\boldsymbol{K}$, $\boldsymbol{V}$, $\boldsymbol{Q}$ projections are computed by learnable transformation matrices on $\boldsymbol{(F+P)}$, where $\boldsymbol{F}$ is an extracted embedding matrix, and $\boldsymbol{P}$ is a positional encoding matrix. Later, reweighted embeddings computed as:

\begin{equation}
\label{self_attn_form}
RHead = Softmax(\frac{QK^T}{\sqrt d})V
\end{equation}
\begin{align*}
RMultiHead = Concat(RHead_0,RHead_1,\\ \dots,RHead_n)W_o
\end{align*}
where $d$ - is the dimensionality of a feature vector, $RHead$ - reweighted representation received from single attention head,  $RMultiHead$ - final reweighted representation  from multiple heads  and $W_o$ - output transformation matrix. Each reweighted projection is assigned an element-wise or component-wise score (as in \cite{yang2017neural} or \cite{gong2019video}). According to those scores, aggregation is done on the original features in the following way:
\begin{align*}
\boldsymbol{s} &= RMultiHead \cdot W_q &
\boldsymbol{q} &= Softmax(\boldsymbol{s}) \\
\boldsymbol{r} &= \sum_{i=1}^{n} f_i*\boldsymbol{q_i}
\end{align*}
where $\boldsymbol{s}$ is a vector of unnormalized quality scores,  $\boldsymbol{q}$ is a vector of softmax normalized scores,  and  $W_q$ is a learnable matrix to retrieve  the quality score of given projection.

\begin{figure}
   \begin{center}
   \label{fig:multi_backbone}
   \includegraphics[width=1.0\linewidth]{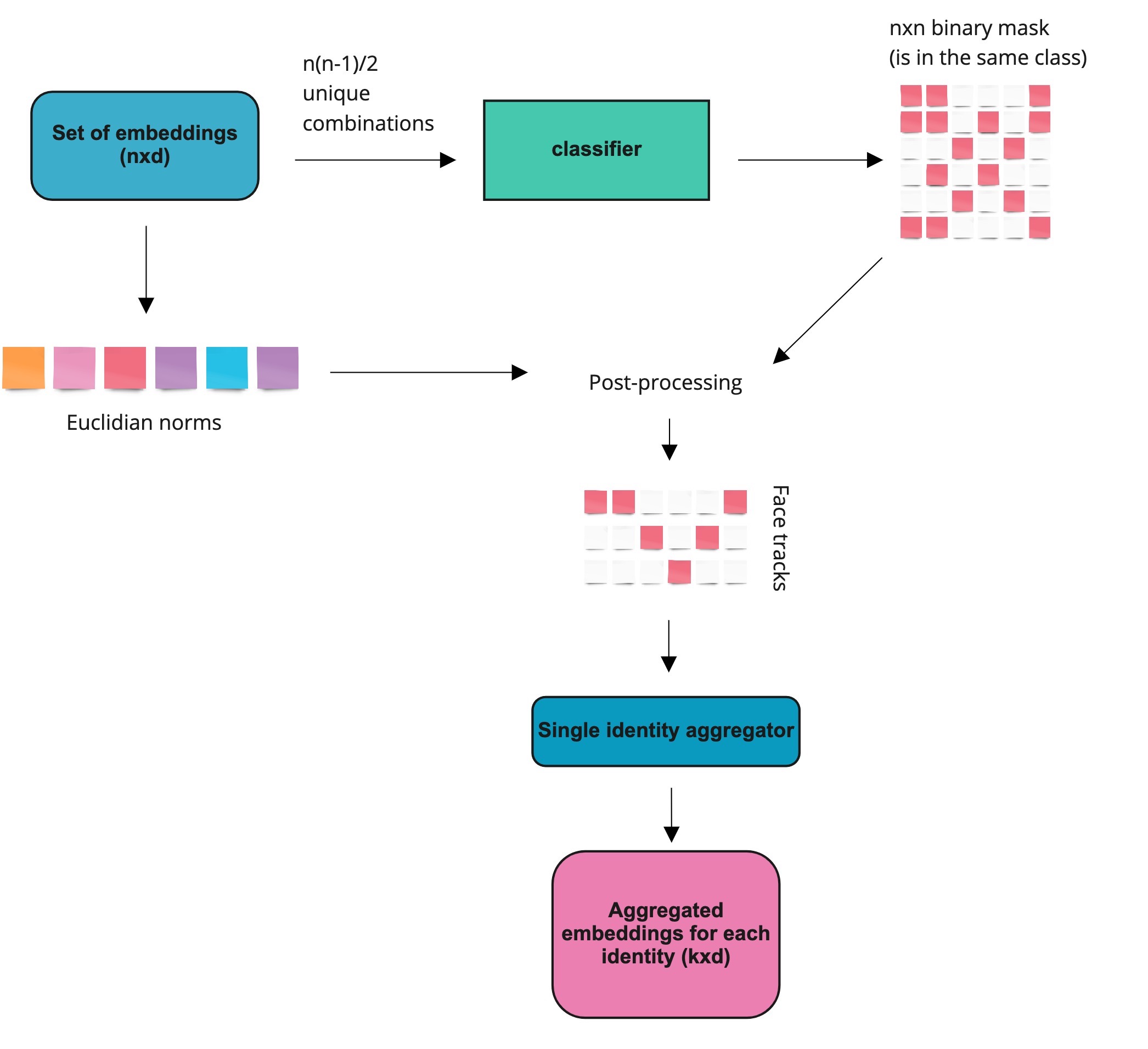}
   \end{center}
   \caption{Proposed multi-identity architecture. $n$ - specifies the number of embeddings in a video, $d$ - dimensionality of the embedding, $k$ - number of unique identities within a video.}
\end{figure}

\subsection{Multi-identity aggregation}
The formulation of the task is similar to section \ref{sident} with the output being a set of vectors $[r_1,r_2,\dots,r_k]$ where $k$ is the number of unique identities within a video sequence. Given a set of $n$ features, multiple identities aggregation could be decomposed into two parts:
\begin{enumerate}
  \item finding $k$ face tracks within a set.
  \item aggregating features withing each face track.
\end{enumerate}

By representing relation to the face tracks as a $(n,n)$ binary matrix, our goal is to find a set of cliques. First, to create a binary mask, we need to classify all possible feature pairs within a video (that is ${n}\choose{2}$). After that, to remove any ambiguity and create a set of face tracks, we need to separate features into strongly connected components. To do that we implement a greedy post-processing. According to \cite{ranjan2018crystal} norm of the embedding could be considered as a proxy to the quality of face embedding. Assignment of components is performed in descending order specified by the Euclidian norm of the embeddings:  $norm_i= ||X_i||_2$, where $X$ is a matrix of extracted embeddings. All elements which lie in a class relation with a given example $i$  are retreived (e.i $\forall j\in {\mathbf{R}(X_i,X_j)}$, where $\mathbf{R}$ is a class relation), assigned to the component and zeroed out in the initial binary matrix. The code snippet on Python is below.

\begin{lstlisting}[language=Python]
import numpy as np

def postprocess_mask(p_mask, q_scores):
    '''
    p_mask :np.array(nxn)  predicted  mask
    q_scores : np.array(nx1) l2 norms of input 
               embeddings
    '''
    p_mask_c = p_mask.copy()
    f_mask = np.zeros_like(p_mask)
    q_sorted = np.argsort(q_scores)
    
    for q_ind in q_sorted[::-1]:
        f_mask[q_ind] = p_mask_c[q_ind]
        non_zero_els = np.argwhere(
                     f_mask[q_ind]!=0
                     )
        p_mask_c[:,non_zero_els] = 0
        p_mask_c[non_zero_els,:] = 0
        
    return f_mask
    
\end{lstlisting}
\label{greedy_postprocessing}
Starting from the embedding with the highest norm, cliques are greedily assigned, until no elements are left unassigned. The resulting matrix is converted to the a of zero-padded face tracks and then aggregated using the single identity aggregation network. 

\section{Experiments}

\subsection{Datasets and protocols}

\begin{figure}
\begin{center}
   \includegraphics[width=1.0\linewidth]{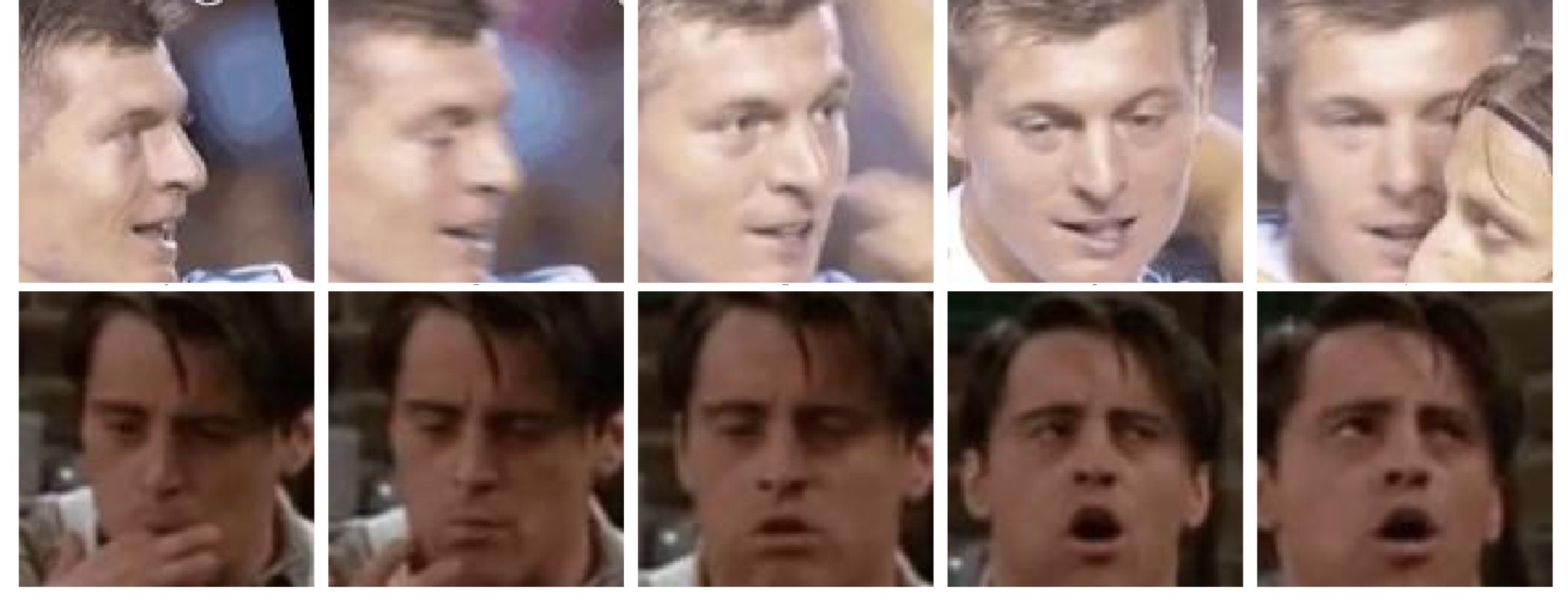}
   \end{center}
   \caption{Face tracks sampled from GIFs dataset.}
   \label{fig:gifs_examples}
\end{figure}

To train our aggregator, we use the \textbf{UMDFaces} \cite{bansal2016umd} dataset. Additionally, we append a dataset with short GIFs of different identities parsed from publicly available API through \href{https://giphy.com/}{giphy.com}. Collected and preprocessed dataset will be made publicly available. For GIFs collection, we used their public API to get the most relevant GIFs for approximately 2300 celebrities. From each GIF, we sampled 5 frames evenly distributed in time. As a result, on average, we obtained about 150 GIFs for each indentity, in total, almost 3M frames. Sampled face tracks and overall training datasets statistics could observed in Figure \ref{fig:gifs_examples} and Table \ref{data_stats}

\begin{table}
\begin{center}
\begin{tabular}{| c | c | c | c | c |}
\hline
\textbf{Dataset} & \multicolumn{1}{ c|}{\textbf{Frames}} & {\textbf{Identities}} & {\textbf{Sessions}}& {\textbf{Length}}   \\ 
\hline 
UMDFaces & 3735476 & 3107 & 22075 & 161\\ \hline
GIFs & 637939 & 2304 & 137739 & 4.6\\ \hline
\end{tabular}
\label{data_stats}
\end{center}
\caption{Training datasets descriptions. Length specified as an average number of frames per session. GIFs statistics correspond to the post-processed dataset.}
\end{table}

\textbf{IJB-C} \cite{maze2018iarpa} dataset is used for  benchmarking. IJB-C has 3,531 indentities with 31.3K still images and 117.5K frames from 11, 779 videos. 

\subsection{Preprocessing}
In the UMDFaces dataset, faces are detected using publicly available MTCNN model \cite{zhang2016joint}. Using 5 facial keypoints, the similarity transform is applied to align faces. To match all detected faces with UMDFaces annotations, an IOU of 0.4 is chosen as a threshold. All non matched  faces are ignored. We use the same detector for GIFs dataset. It is important to note, that since GIF may contain multiple faces and not just the ones of a desired celebrity, we used pretrained ArcFace model to get feature vectors for detected faces, computed mean vectors for each class and filtered out those faces whose vectors do not lie within specified cosine distance threshold of 0.7 from the corresponding class center.

\subsection{Training details}
\label{single_train}
\textbf{Single identity training}:   
The aggregator is trained on the UMDFaces+GIFs dataset using an additive angular margin loss \cite{wang2018additive}. The radius of the hypersphere and additive margin are set to 16.0 and 0.35, respectively. The paramaters are trained using a RAdam optimizer \cite{liu2019variance} with default values. For the encoder, we use an official implementation of \href{https://github.com/tensorflow/models/blob/master/official/transformer/model/transformer.py}{Transformer} on Tensorflow, with a single modification (removing the trainable embedding block). The depth of the encoder is set to 4, the number of heads for the aggregator on top of ResNet and MobileNet is set to 8 and 4, respectively. Attention and ReLU dropout rates are specified as 0.3 and 0.4. All other hyperparameters are populated with the default values. 

For validation, we employ an identity-based split. Due to the fact that identities in the validation are not present in the training set, we define a metric for the early stopping - intra-class proximity gain, defined as:
$$ ICPG = E(Intra(AVE(X))) - E(Intra(SA(X))  $$
where $X$ - is the set of templates for aggregation, $SA$ and $AVE$ are self-attention pooling and average pooling respectively, $Intra(Y)$ - is a set of distances between the aggregated elements of $Y$ which belong to the same class. Specifically, each mini-batch includes 256 templates
that are randomly sampled from 128 identities, 32 images per template. 
\newline

\textbf{Multi-identity training}:    
There are no multi-identity annotated video datasets  to our knowledge, so we synthetically created one (combining UMDFaces and GIFs). We sample a random number of identities (from 2 to 64) and choose a single session for each identity. These sessions are concatenated into a single one, which is then used for sampling 256 frames. These frames form a multi-identity video. The order of frames is preserved and used later during face tracks creation. 

For the binary mask generation, there may be  a trainable classifier, but we have found that simple thresholding based on the cosine distance results in higher recall and precision. To train the identity aggregator on top of the post-processed mask, we employ a teacher forcing with a scheduled sampling \cite{bengio2015scheduled}. For the first 5000 iterations, cosine decay scheduler is used to specify the probability of mixing ground truth masks with the post-processed one. After that, only a predicted mask is used for the aggregation.  

\begin{figure}
   \begin{center}
   \includegraphics[width=1.0\linewidth]{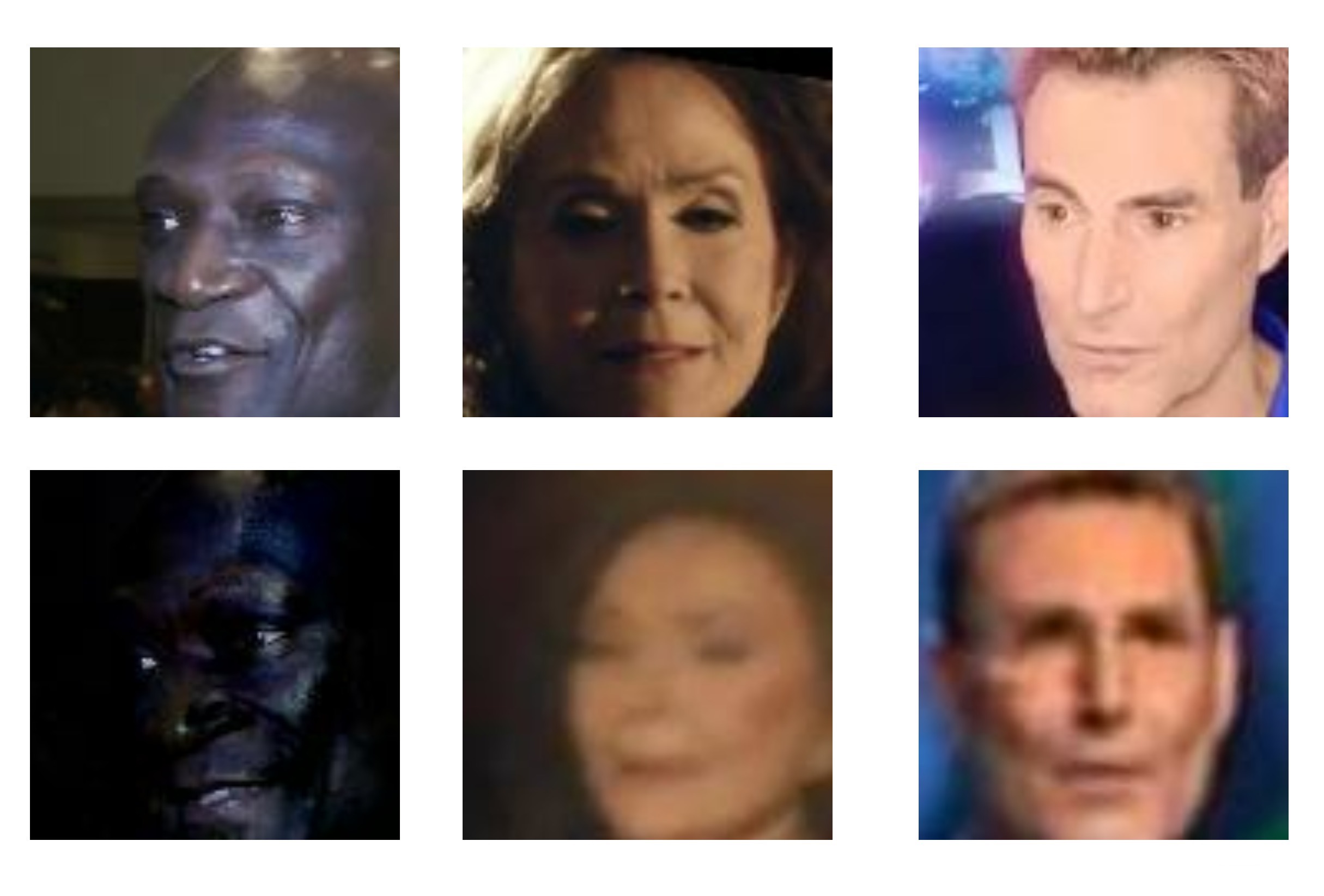}
   \end{center}
   \caption{Attention extremums within the face tracks on the validation set. The first row represents the highest values, the second one - the lowest.}
   \label{fig:attention_dist}
\end{figure}

\subsection{Results for single-identity aggregator}

\subsubsection{UMD+GIFs validation set}


\begin{table}
\begin{center}
\begin{tabular}{| c | c | c | c | c |}
\hline
\textbf{Aggregator} & \multicolumn{1}{ c |}{\textbf{ICPG}} \\
\hline 
SA pooling (ResNet) & 2.8* $1e^-3$  \\ \hline
SA pooling (MobileNet) & 5.5* $1e^-3$ \\ \hline
\end{tabular}
\label{sing_icpg}
\end{center}
\caption{ICPG on the validation set}
\end{table}

\begin{figure}
   \begin{center}
   \includegraphics[width=1\linewidth]{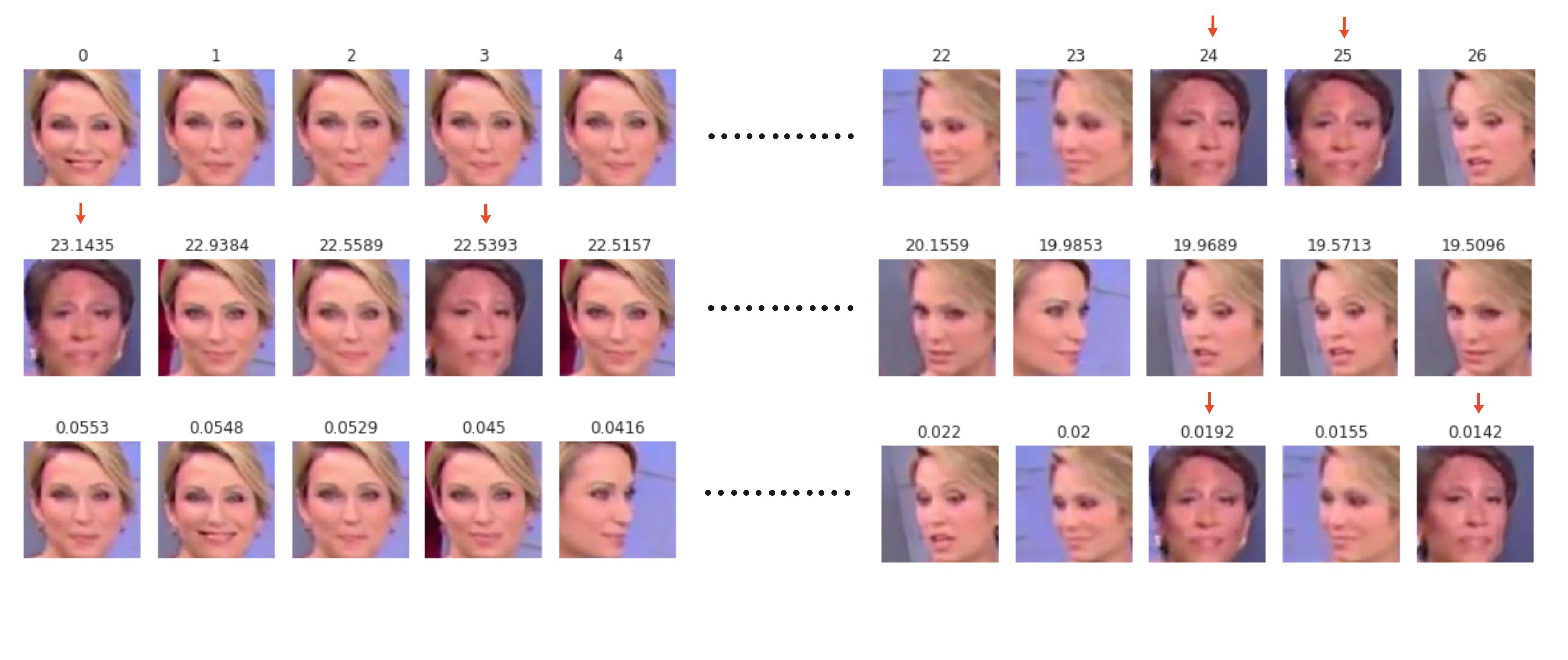}
   \end{center}
   \caption{Attention distribution on the face track with erroneous elements. The first row - ordered frames of the given face track. The second row - frames ordered by the Euclidian norm of their corresponding embedding. The third row - frames ordered according to their self-attention quality score. The elements which do not belong to the given identity are emphasized with the red arrow.}
   \label{fig:attention_dist_error}
\end{figure}

On the validation set, we report ICPG for  the aggregators on top of different feature extractors as show in Table \ref{sing_icpg}. Additionally, we attach visualization of attention extremums on the given set in Figure \ref{fig:attention_dist}.

We have found that there are some face tracks in the UMDFaces which contain erroneously assigned elements. Also, we demonstrate in Figure \ref{fig:attention_dist_error} that by resorting to the general context, the self-attention mechanism helps to downweight such frames during final aggregation (despite their high visual quality).

\subsubsection{IJB-C}
1:1 Verification on mixed set and 1:N end-to-end video probes are tested. AVE (average) and SA (self-attention) poolings are compared and results shown in Table \ref{sing_roc}. 

The proposed approach results in a higher AUC, TAR, top-1 accuracy for both aggregators based on MobileNet and ResNet feature extractors. Notably, we observe that there is a more significant gap between aggregation and averaging based on the MobileNet feature extractor. It shows that the aggregation is more beneficial on the networks with the lower representational capacity. 

\begin{table}
\begin{center}
\begin{tabular}{| c | c | c | c | c |}
\hline
\textbf{Aggregator} & \multicolumn{2}{ c |}{\textbf{Ver}} & {\textbf{Id}}   \\ 
\cline{2-4} & \textbf{TAR} & \textbf{AUC} &  \textbf{Rank-1}\\
\hline 
AVE pooling (ResNet) & 83.90 & 99.39 & 82.54  \\ \hline
SA pooling (ResNet) & \textbf{87.83} & \textbf{99.44} & \textbf{82.79}\\ \hline
AVE pooling (MobileNet) & 55.58 & 98.92 & 66.73  \\ \hline
SA pooling (MobileNet) & \textbf{62.04} & \textbf{99.01} & \textbf{67.48}\\ \hline
\end{tabular}
\label{sing_roc}
\end{center}
\caption{Verification TAR is reported under FAR=$1e-6$, identification rate is reported as rank-1 accuracy.}
\end{table}

\begin{figure}
   \begin{center}
   \label{fig:roc_resnet_agg}
   \includegraphics[width=1.0\linewidth]{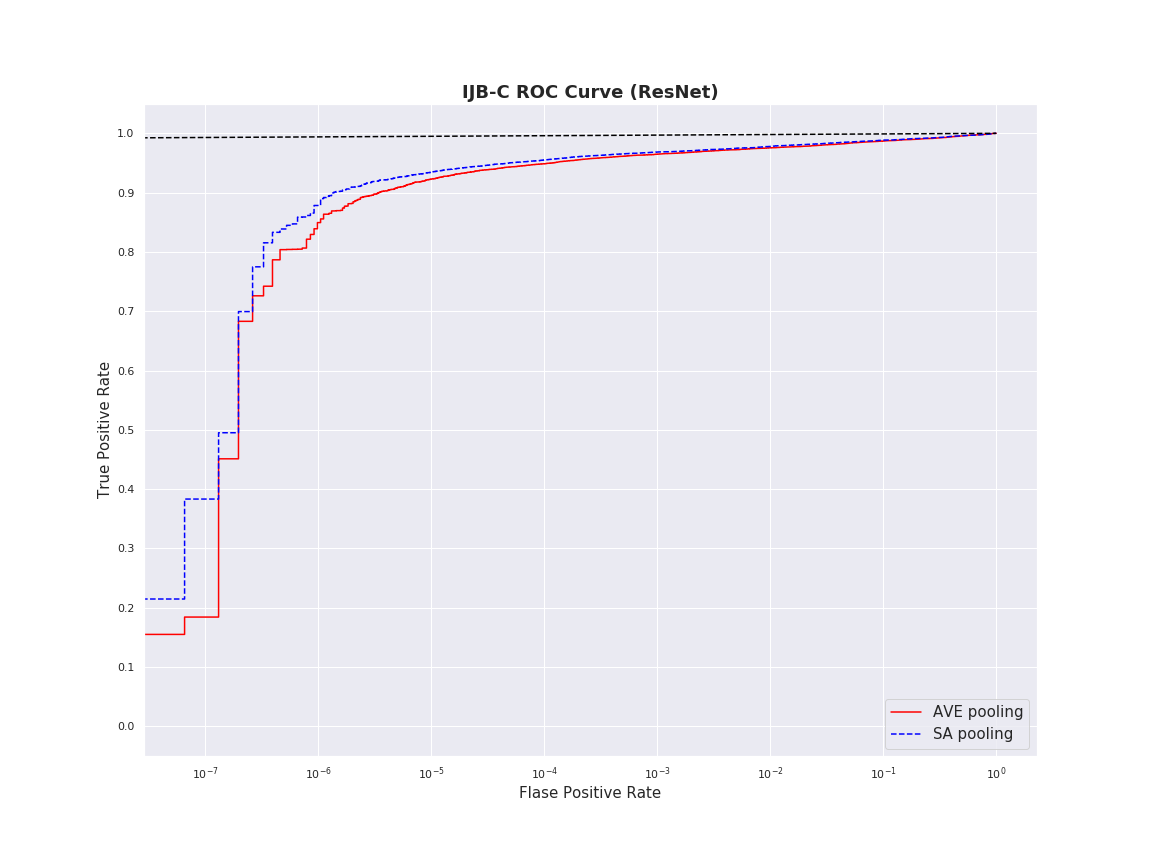}
   \includegraphics[width=1.0\linewidth]{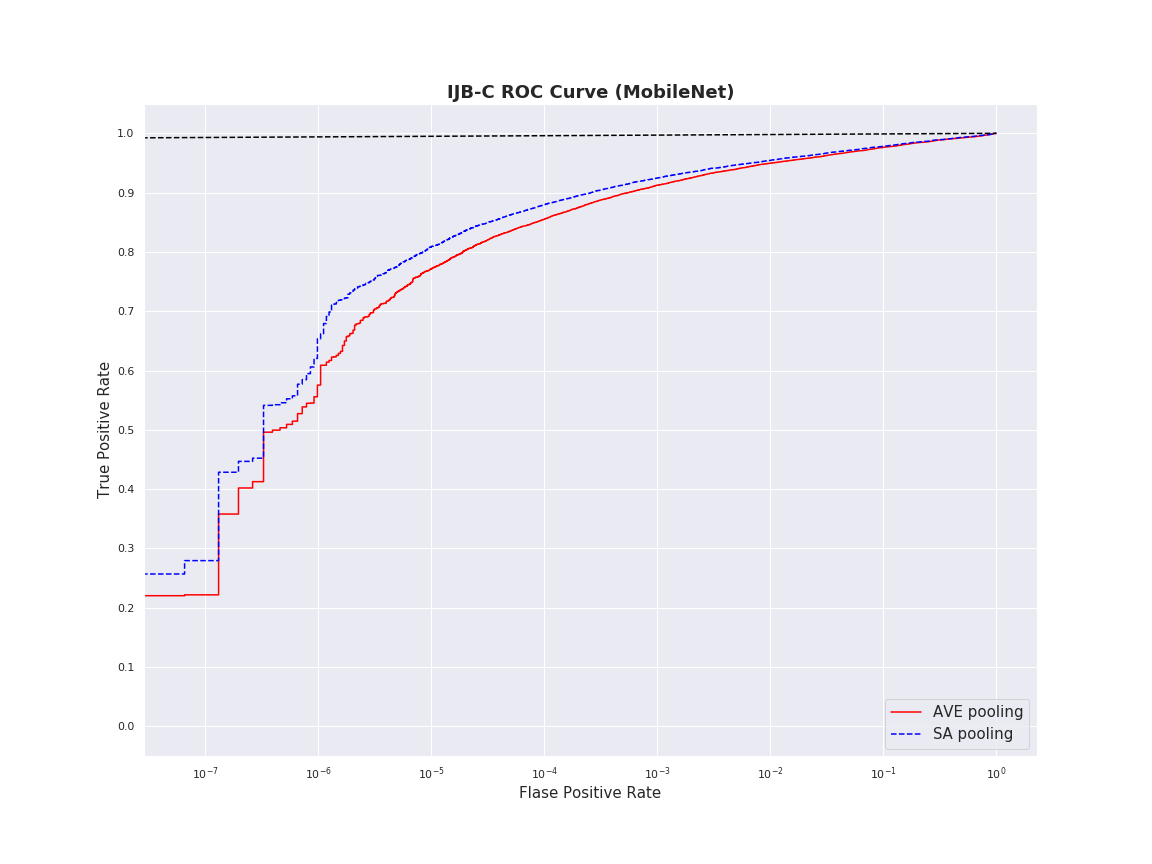}
   \end{center}
   \caption{ROC curves of SA Pooling vs AVE Pooling on ResNet (top) and MobileNet (bottom) feature extractors.}
\end{figure}

\subsection{Results for multi-identity}
\subsubsection{UMD+GIFs validation set}

The number of regressed identities is calculated on the proposed multi-identity videos. Above mentioned binary mask producer results in MPE (mean percentage error) of $4.2\%$. We presume that using different post-processing techniques for masks, such as transitive closure, could be used to lower the MPE.

\subsubsection{IJB-C}

\begin{figure}
   \begin{center}
   \label{fig:roc_resnet}
   \includegraphics[width=1.0\linewidth]{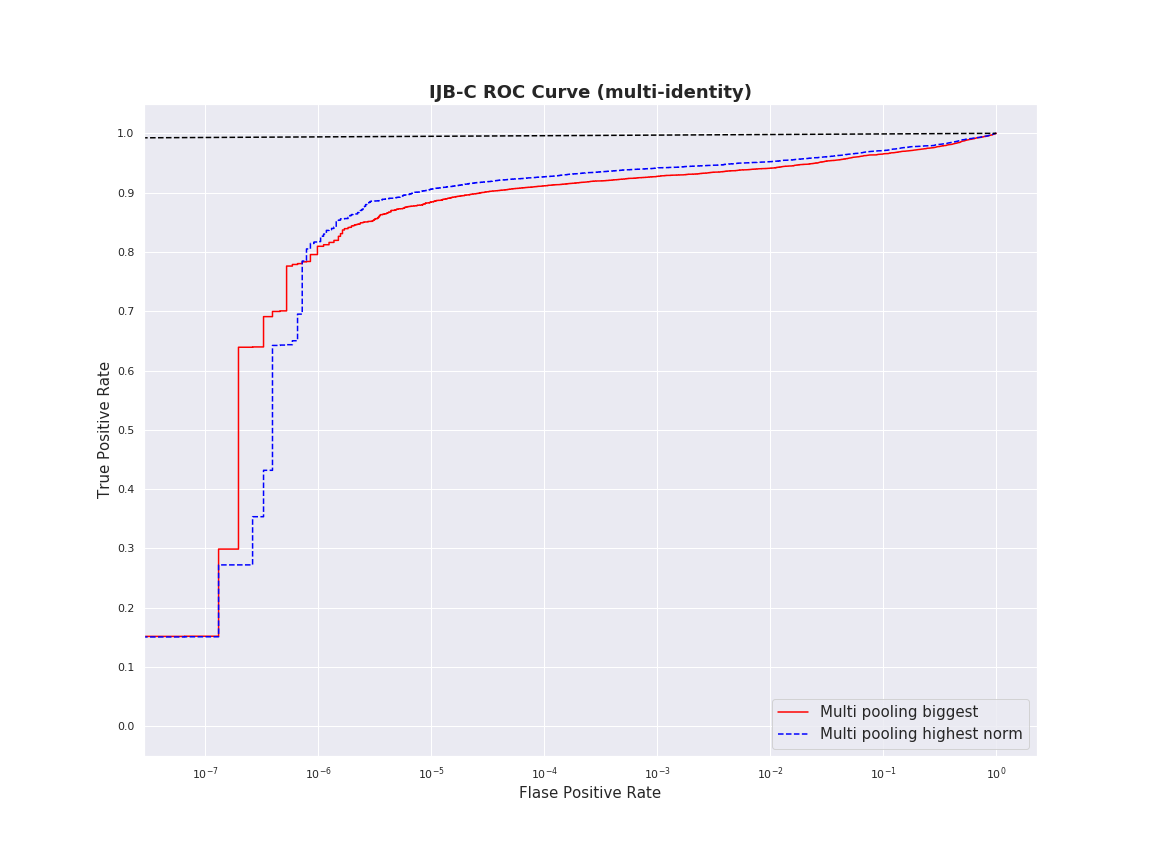}
   \end{center}
   \caption{ROC curves of multi-identity aggregator with the highest norm and the biggest component selection.}
   \label{fig:roc_multi}
\end{figure}

\begin{table}
\begin{center}
\begin{tabular}{| c | c | c | c | c |}
\hline
\textbf{Multi-identity aggregator} & \multicolumn{2}{ c |}{\textbf{Ver}} & {\textbf{Id}}   \\ 
\cline{2-4} & \textbf{TAR} & \textbf{AUC} &  \textbf{Rank-1}\\
\hline 
Highest norm component  & \textbf{81.67} & \textbf{98.58} & \textbf{82.54}  \\ \hline
Biggest component & 79.55 & 98.30 & 82.02\\ \hline
\end{tabular}
\label{multi_ijb}
\end{center}
\caption{Verification TAR is reported under FAR=$1e-6$, identification rate is reported as rank-1 accuracy. Different component selection strategies are compared.} 
\end{table}

Having only a single identity within a session, we introduce two component selection strategies (in order to select a single representation for reference in the session where multiple identities were detected). Component with largest number of frames and a component which contains the embedding of the highest quality are tested. Results are reported in Table \ref{multi_ijb} and ROC curve is displayed on Figure \ref{fig:roc_multi}.

\section{Conclusions}

In this paper, we propose a novel self-attention aggregation network for learning face representation for any number of identities from a video stream. We show that SAAN outperforms average pooling in general for a single identity. Especially, the usage of SAAN network could be beneficial if vector representation quality degrades, e.g. when using lightweight embeddings from MobileNet. Moreover, the further investigation indicates that SAAN model is robust to erroneous face tracks. Also, we created a dataset for multi-identity aggregation problem and plan to make it available under MIT licence. Our future work will explore the different mask postprocessing approaches and ways to improve the aggregation model for multi-indentity video streams.


{\small
\bibliographystyle{ieee_fullname}
\bibliography{egpaper_final}
}

\end{document}